\documentclass{article}
\usepackage{ijcai24}

\usepackage{times}
\usepackage{soul}
\usepackage{url}
\usepackage[hidelinks]{hyperref}
\usepackage[utf8]{inputenc}
\usepackage[small]{caption}
\usepackage{graphicx}
\usepackage{amsmath}
\usepackage{amssymb}
\usepackage{amsthm}
\usepackage{booktabs}
\usepackage{multirow}
\usepackage{algorithm}
\usepackage{algorithmic}
\usepackage[switch]{lineno}

\usepackage{color}
\usepackage{xcolor}

\title{A General Black-box Adversarial Attack on Graph-based Fake News Detectors}

\author{
Peican Zhu$^1$
\and
Zechen Pan$^2$\and
Yang Liu$^{1*}$\and
Jiwei Tian$^3$\and
Keke Tang$^{4*}$\And
Zhen Wang$^{1}$\footnote{Corresponding authors.}\\
\affiliations
$^1$School of Artificial Intelligence, Optics and Electronics, Northwestern Polytechnical University\\
$^2$School of Computer Science, Northwestern Polytechnical University\\
$^3$Air Traffic Control and Navigation College, Air Force
Engineering University\\
$^4$Cyberspace Institute of Advanced Technology, Guangzhou University\\
\emails
\{ericcan, w-zhen\}@nwpu.edu.cn,
928598047@mail.nwpu.edu.cn,
\{yangliuyh, tangbohutbh\}@gmail.com,
tianjiwei2016@163.com
}

\begin{document}

\maketitle

\begin{abstract}
Graph Neural Network (GNN)-based fake news detectors apply various methods to construct graphs, aiming to learn distinctive news embeddings for classification. Since the construction details are unknown for attackers in a black-box scenario, it is unrealistic to conduct the classical adversarial attacks that require a specific adjacency matrix. In this paper, we propose the first general black-box adversarial attack framework, i.e., General Attack via Fake Social Interaction (GAFSI), against detectors based on different graph structures. Specifically, as sharing is an important social interaction for GNN-based fake news detectors to construct the graph, we simulate sharing behaviors to fool the detectors. Firstly, we propose a fraudster selection module to select engaged users leveraging local and global information. In addition, a post injection module guides the selected users to create shared relations by sending posts. The sharing records will be added to the social context, leading to a general attack against different detectors. Experimental results on empirical datasets demonstrate the effectiveness of GAFSI.
\end{abstract}

\section{Introduction}

\begin{figure}[!tb]
    \centering
    \includegraphics[width=0.98\linewidth]{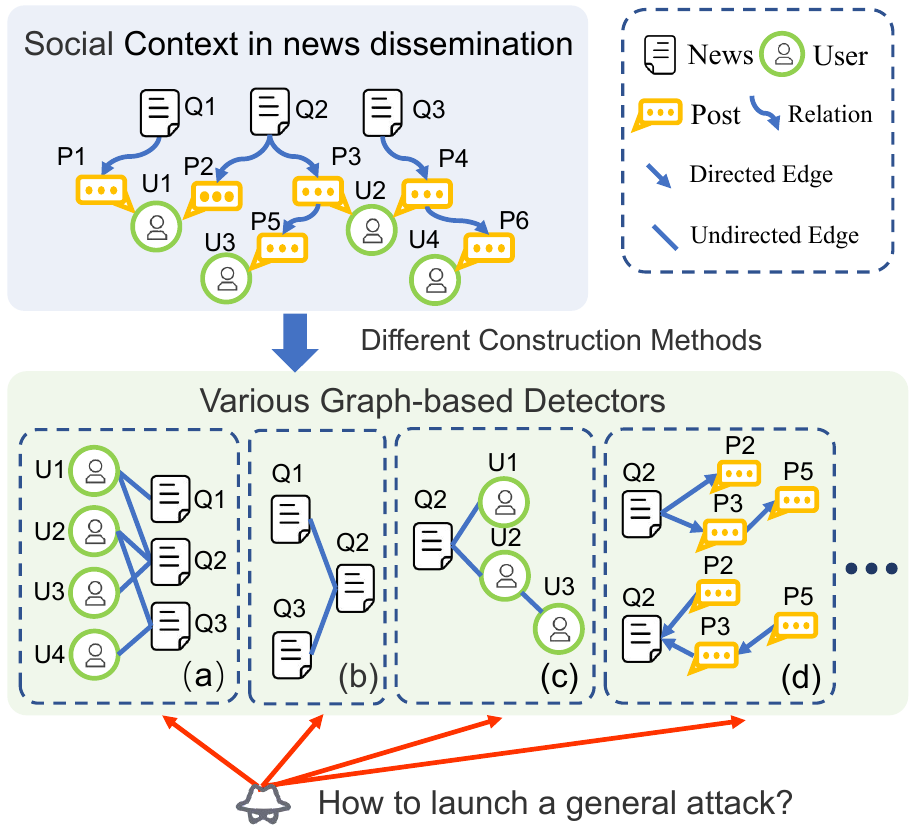}
    \vspace{-2mm}
    \caption{
    Social context in news dissemination can be constructed into (a) the user-news bipartite graph; (b) the news engagement graph; (c) the news-user propagation tree; and (d) the news-post propagation and dispersion tree, etc.
 The diversity in the graph types poses challenges for launching a general black-box adversarial attack on fake news detectors based on different graphs.}
    \label{fig_atkdiff}
\end{figure}

The rapid expansion of online social media platforms has led to a rise in misinformation, undermining public trust in truth and science. Unlike traditional media, where content is often rigorously fact-checked, the interactive nature of social media accelerates the spread of fake news through commenting and sharing, magnifying its impact. This makes detecting and countering fake news on these platforms both challenging and crucial \cite{Aïmeur:fnreview2023,Cheng:ginsd2024}.

To tackle this challenge, abundant machine-learning methods are proposed for fake news detection. Besides utilizing single-modal or multi-modal detectors to extract features from content \cite{Dong:algm2023,Hua:ttec2023,Dong:nslm2024}, there is a growing interest in conceptualizing the online social network as a graph structure to leverage the rich social context \cite{Phan:asc2023,Yin:gamc2024}. Among existing studies, the user-news bipartite graph is commonly used for detectors to model user engagement \cite{Nguyen:fang2020,Wang:marl2023,Su:usdefake2023}. Besides, the news engagement graph can be constructed to explore relations between news directly \cite{Wu:decor2023}. Instead of capturing information in the global network, the news's local propagation structure is also investigated. The news-post propagation and dispersion tree \cite{Bian:bigcn2020} are constructed to aggregate information from two directions, while the news-user propagation tree \cite{Dou:upfd2021} is constructed to capture user-aware in the propagation structure. Due to the diversity of the graph construction, Graph Neural Network (GNN) is allowed to learn distinctive news embeddings from different perspectives for news classification. Despite their effectiveness in detecting fake news, these detectors have increasingly been found to be vulnerable to adversarial attacks \cite{Wang:marl2023}. This vulnerability may be utilized to manipulate public opinion or gain financial benefits. Therefore, it is critical to investigate adversarial attacks on GNN-based fake news detectors to assess and enhance their robustness.

Existing adversarial attacks on GNN are primarily categorized into two types: edge perturbation and node injection. Edge perturbation techniques aim to alter the model's prediction by adding or removing edges in the graph~\cite{Zügner:nettack2018,Li:sga2021,Zhang:mibtack2023}, while node injection approaches focus on generating and introducing malicious nodes into the graph~\cite{Tao:gnia2021,Wang:clusterattack2022}. Although these methods show promise when the target graph is exposed, their efficacy diminishes in a more realistic black-box setting where the graph construction process is unknown. Given the vast differences in fake news detectors based on various social network structures, assuming a certain graph to conduct attacks becomes impractical for attackers. This situation leads to a crucial question: Is it possible to develop a universal attack strategy that remains effective in a black-box scenario against fake news detectors based on diverse social network structures?

In this paper, we propose a General Attack via Fake Social Interaction (GAFSI) against GNN-based fake news detectors, aiming to alter the prediction of the target news. Unlike existing studies, that are concentrated on a certain graph, we attempt to add social interaction records into the social context and thus can influence the graph representation regardless of the construction methods. 
Specifically, we employ a fraudster selection module to identify influential fraudsters for creating fraudster-post pairs. Then, through a post injection module, we select the source of new posts in the propagation structure and clone content from existing posts to new ones. Accordingly, the selected fraudsters will send the posts to simulate social interactions. This integration of social interaction records can effectively perturb the social context and fool GNN-based fake news detectors. We validate our framework's effectiveness in attacking various GNN-based fake news detectors with different graph structures. Our extensive experimental results demonstrate that GAFSI is general and capable of deceiving different detectors in a black-box setting, outperforming state-of-the-art methods.

Overall, our contribution is summarized as follows:
\begin{itemize}
\item We propose a general black-box attack framework to simulate social interactions, enhancing the effectiveness of the perturbation in a more realistic scenario. 


\item We propose GAFSI which leveraging gradient-based attention information, consists of a fraudster selection module and a post injection module.

\item Extensive experimental results demonstrate that the perturbation generated by GAFSI on social context can effectively change the prediction of the target news.

\end{itemize}

\section{Related Work}

\subsection{GNN-based Fake News Detection}
According to the graph prototypes, GNN-based fake news detectors are primarily categorized into two types: propagation-based detectors and social-context-based detectors \cite{Phan:asc2023}. The propagation-based detectors employ a tree-structured graph to model the propagation process of a single news \cite{Bian:bigcn2020,Dou:upfd2021,Silva:p2v2021,Han:clfn2021}. Here, the root node represents a news article and other nodes represent associated posts, while edges depict shared relations among nodes. Besides, social-context-based detectors prefer to capture global information. Recent work \cite{Nguyen:fang2020} considers the relations among publishers, news, and users to construct the social context graph. Su \emph{et al.} \shortcite{Su:usdefake2023} construct a dual-layer graph that contains a news propagation layer and a user interaction layer. They use shared relations to connect the two layers. Wu and Hooi \shortcite{Wu:decor2023} construct a news engagement graph based on the number of common users. Although some methods utilize additional information to construct the graph, it is worth noting that shared relation remains a key factor in those graph construction processes. Hence, it inspired us to fool the GNN-based fake news detector via fake social interaction.

\subsection{Adversarial Attack on GNN}
Numerous adversarial attack methods have been proposed to fool classifiers in various domains \cite{Zhu:sgma2023,Tang:iot2023,Zhu:ham2024,Guo:mixcam2024}. For graph-related tasks, the attack techniques include attribute perturbation, edge perturbation, and node injection \cite{Liu:grad2022}. Zügner \emph{et al.} \shortcite{Zügner:nettack2018} first conduct targeted attacks on GNN through greedy-based edge perturbation. To address the inefficiency of the previous methods on large-scale graphs, Li \emph{et al.} \shortcite{Li:sga2021} introduce a method containing a subgraph constructing process. Zhang \emph{et al.} \shortcite{Zhang:mibtack2023} focus on finding a minimized set of perturbations to realize an attack with low cost. Recent work \cite{Shang:ea2023} start to conduct edge perturbation on heterogeneous graphs. They consider that all GNNs take the same graph as input and focus on perturbing the edge that has a larger common weight. Instead of edge perturbation, Tao \emph{et al.} \shortcite{Tao:gnia2021} propose a targeted attack method by injecting a single node which is obtained from a pre-trained encoder. Wang \emph{et al.} \shortcite{Wang:clusterattack2022} regard the node injection attack as a graph clustering problem and solved it based on Euclid’s distance between nodes' adversarial feature. Although existing attacks perform well in traditional graph tasks, they demonstrate limitations when attacking GNN-based fake news detectors. The edge perturbation method may break the chain, leading to the destruction of the propagation structure, which is easy to discover. Furthermore, the node injection method generates perturbation in the form of embedding which can not be restored to original attributes. Hence, they are insufficient to conduct attacks against fake news detectors. 

Wang \emph{et al.} \shortcite{Wang:marl2023} is the first to attack GNN-based fake news detectors, focusing on manipulating three types of fraudsters to collaborate in deceiving the detector. They formulate the attack problem as attacking GNNs on a user-news bipartite graph and successfully alter the predicted label of the target news. However, they only consider modifying the edge between fraudsters and target news in the user-news bipartite graph. Thus, similar to previous attacks, their method can not attack detectors based on other graph structures. 


\section{Preliminary and Problem Statement}

\paragraph{Social Context in News Dissemination.}
The social context in news dissemination is represented as $G=(U, Q, P, E)$, where $U=(u_1,\dots,u_m)$ indicates a set of $m$ users, $Q=(q_1, \dots,q_n)$ denotes a set of $n$ news, $P=(p_1, \dots,p_l)$ represents a set of $l$ posts, $E=\{E_u,E_t\}$ represents a set of relation in $G$. Here, $E_u$ characterizes the user-post relations, with an edge $(u_i,p_j)\in E_u$ signifying that user $u_i$ is the author of post $p_j$. $E_t$ characterizes a set of the shared relations between news and posts, as well as the shared relations among posts themselves. Then, the fake news detection task based on $G$ can be treated as a binary classification problem. $y_i\in Y$ is the ground-truth label of the news $i$ where 0 and 1 represent the real news and fake ones respectively.

\paragraph{Graph Construction of $G$ for Fake News Detection.}
To mine rich information in social context $G$, GNN-based fake news detectors typically consider the entities and relations that can represent a kind of structural pattern to construct a graph. As there are abundant structural patterns in the social context, the graph constructed by different detectors encompasses a variety of types, e.g., the user-news bipartite graph~\cite{Wang:marl2023}, the news engagement graph~\cite{Wu:decor2023}, the news-user propagation tree~\cite{Dou:upfd2021}, and the news-post propagation and dispersion tree~\cite{Bian:bigcn2020}. For simplicity, we describe the graph construction process as:
\begin{equation}
\begin{aligned}
G_s = r(G),
\end{aligned}
\label{eq_gs}
\end{equation}
where $r(\cdot)$ is a function to generate a graph based on social context. After constructing the graph, detectors will utilize a pre-trained language model, e.g., GloVe \cite{Pennington:glove2014} and BERT \cite{Devlin:bert2019}, to generate the feature matrix $X$ and employ a GNN to learn news embedding, aiming to classify news.

\paragraph{Adversarial Attack on Specific Graph.}
Existing adversarial attack methods can easily perturb the graph by adding a small perturbation $\xi$ to the graph $G_s$. Here, the attack process can be described as:
\begin{equation}
\begin{aligned}
G^\prime_s = G_s + \xi.
\end{aligned}
\label{eq_gsprime}
\end{equation}
However, different detectors employ different functions to generate the graph, and thus attackers need to deal with a set of Graphs $\{G_s^1, \dots, G_s^k\}$ where $k$ is the number of the detectors. Due to structural differences, perturbation $\xi$ is not general for the graphs in the set. In a black-box manner, the problem will become more challenging, thereby encouraging us to develop a general approach to attack the detectors.

\paragraph{Our Attack on Social Context.}
To execute a generalized attack against GNN-based fake news detectors, a natural way is to perturb the social context $G$ instead of $G_s$. Therefore, a general attack process can be described as:
\begin{equation}
\begin{gathered}
G^\prime = G + \xi.
\end{gathered}
\label{eq_gprimexi}
\end{equation}
However, the small perturbation $\xi$ may disrupt the semantic integrity, making attacks more easily detectable. Hence, we attempt to simulate social interaction to meet practical constraints. Specifically, we manipulate a set of controllable fraudsters $U_c$ to add social interaction records through sharing. The perturbed social context can be represented as:
\begin{equation}
\begin{gathered}
G^\prime=(U, Q, P \cup P^\prime, E\cup E^\prime_u \cup E^\prime_t),
\end{gathered}
\label{eq_gprime}
\end{equation}
where $P^\prime$ includes new posts sent by fraudsters, $E^\prime_u$ and $E^\prime_t$ are new user-post and shared relations, respectively. The attack's objective for the $k$-th news $q_k$ is formalized as:
\begin{equation}
\begin{gathered}
\mathop{\max}\limits_{P^\prime,E^\prime}\mathbb{I}(\mathop{\arg\max}\limits_{z}f_{\theta^*}(r(G^\prime), X)_{q_k}\neq y_{q_k}), \\
\text{s.t.}\ \theta^*=\mathop{\arg\min}\limits_{\theta}L(\theta;r(G),X), \\
|P^\prime|=|E^\prime_u|=|E^\prime_t|\leq \Delta ,
\end{gathered}
\label{eq_problem}
\end{equation}
where $f_\theta$ is the targeted GNN-based fake new detector which has been trained on the clean data, $\mathbb{I}(\cdot)$ is an indicator function, $X$ is the feature matrix, $z$ is the predicted label of news $q_k$, and $\Delta$ is the given budget. The goal is to maximize the likelihood of $q_k$ being misclassified within the constraints.

\begin{figure*}[tb]
    \centering
    \includegraphics[width=6.9in]{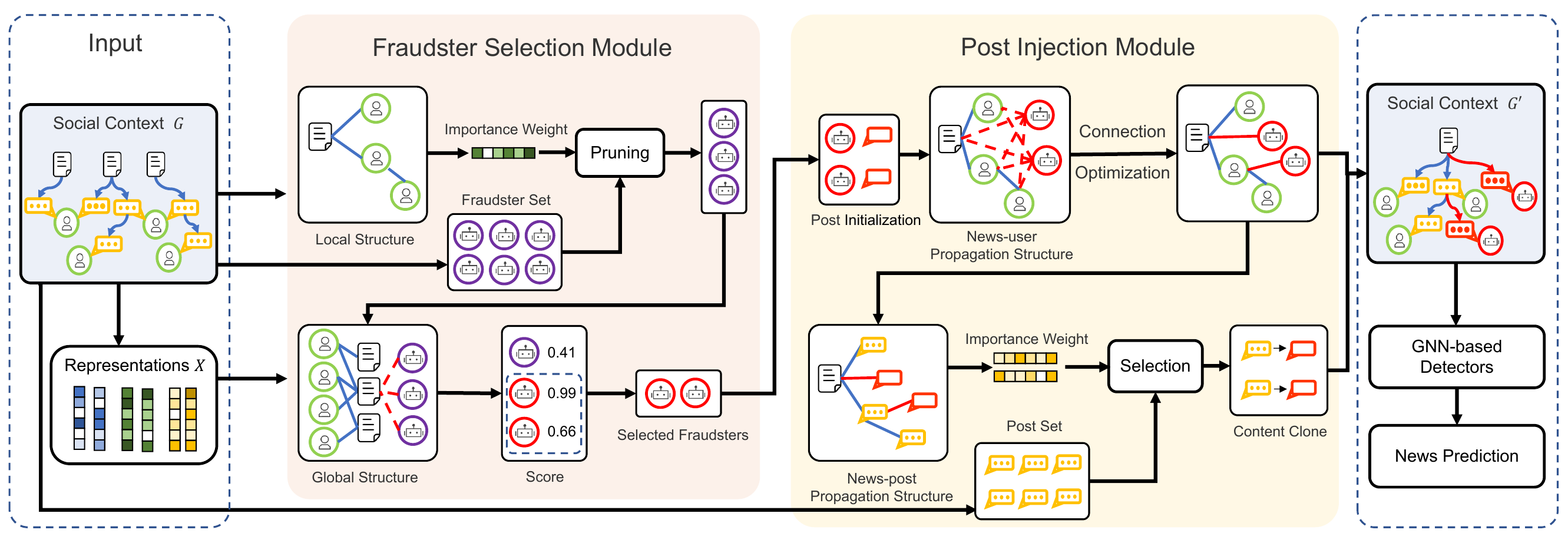}
    \caption{An illustration of GAFSI against GNN-based fake news detectors. The social context $G$ and corresponding representation $X$ serve as the input of GAFSI. The fraudster selection module selects influential fraudsters according to information from the local structure and global structure. Each selected fraudster will send a post and then the post injection module will optimize the connection and the content of the post. Finally, the records of sharing will be added into the social context and fool the GNN-based detector.}
    \label{fig_atk}
\end{figure*}

\section{Method}

In this section, we describe our general attack framework which consists of two major modules, i.e., the fraudster selection module and the post injection module, as shown in Figure \ref{fig_atk}. Given the social context of news dissemination $G$ as input, our framework adds a set of social interaction records and eventually fools the GNN-based detectors.

\subsection{Fraudster Selection Module}
To select users to engage in the attack, following ~\cite{Dou:upfd2021,Wang:marl2023}, we first utilize a pre-trained language model~\cite{Pennington:glove2014} to extract user representation from the users' historical posts. However, the GNN-based detector's attention on the user representation is often different from the local structure and global structure. To comprehensively consider the impact of user representation on different structures, we utilize the propagation tree to estimate the local influence of the user representation and construct the user-news bipartite graph to estimate the global influence. Accordingly, we propose a user selection module that can effectively optimize the set of potential fraudsters.

\paragraph{Local Influence Estimation.}
To capture local influence, we first construct a propagation tree $G^k_t=(\mathcal{V}^k_t,E^k_t)$ for target news $q_k$, which consists of:
\begin{equation}
\begin{aligned}
\mathcal{V}^k_t=&\{v| \forall v\in \text{traversal }(E_t, q_k)\},\\
E^k_t =& \{(v_i, v_j)| \forall (v_i, v_j)\in \text{traversalpath }(E_t, q_k)\} , \\
\end{aligned}
\label{eq_gkt}
\end{equation}
where $\text{traversal}(E_t, q_k)$ indicates a function to traverse a tree with news $q_k$ as the root according to $E_t$, and $\text{traversalpath}(E_t, q_k)$ indicates a function to obtain the paths in above traverse process. Then, we let the representations of post nodes in $G^k_t$ equal to the user representation of the post owner such that the GNN-based detector can learn user representation in a local view. Empirically, the gradient can characterize the model's attention on different inputs. Inspired by gradient-based attribution methods \cite{Shrikumar:deeplift2017,Sanchez-Lengeling:eva2020}, we leverage the model's attention on different feature elements to measure the local influence. Following the black-box manner, a GNN-based graph classifier $f_{\theta_t}$ is trained as a surrogate model to obtain attention information. Given the propagation tree $G^k_t$, the importance weight of the user representation is given by: 
\begin{equation}
\begin{gathered}
w_{G^k_t} = \frac{1}{N-1}\sum_{v_i\in \mathcal{V}^k_t\setminus \{q_k\}}\frac{\partial f_{\theta_t}(A_{G^k_t},X_{G^k_t})_{1-y_{q_k}}}{\partial x_{v_i}},\\
\end{gathered}
\label{eq_sx}
\end{equation}
where $A_{G^k_t}$ is the adjacency matrix of $G^k_t$, $X_{G^k_t}$ is the user representation matrix of $G^k_t$, $x_{v_i}$ is the user representation of the publisher of post $v_i$ and $N$ is the scale of $\mathcal{V}^k_t$. The $j$-th entry in $w_{G^k_t}$ indicates the relative importance for $j$-th feature element resulting from classifying $q_k$ into the wrong class based on $G^k_t$. Intuitively, users with greater influence should have larger weights on the relatively important feature elements. According to the importance weight, the local influence of a fraudster $u_i \in U_c$ is measured as:
\begin{equation}
\begin{gathered}
s_{u_i} = x_{u_i} \cdot w_{G^k_t},\\
\end{gathered}
\label{eq_su}
\end{equation}
where $x_{u_i}$ is the user representation of the fraudster $u_i$. To improve efficiency, we utilize local influence for pruning. Obviously, nodes with lower local influence are not ideal choices. Hence, $ceil(\alpha\Delta)$ fraudsters will be remained, where their local influence are $ceil(\alpha\Delta)$-largest.

\paragraph{Global Influence Estimation.}
To estimate the global influence of the user representation, we construct a user-news bipartite graph $G_b=(\mathcal{V}_b, E_b)$:
\begin{equation}
\begin{aligned}
\mathcal{V}_b=&U\cup Q, \\
E_b =& \{(q_k,u_i)|\exists (u_i,p_j)\in E_u \land p_j\in \mathcal{V}^k_t\}. \\
\end{aligned}
\label{eq_gb}
\end{equation}
Due to the direct connections between users and news, we can utilize the gradient of the edge between them to estimate the global influence. Similarly, we train a GNN-based node classifier $f_{\theta_b}$ as the surrogate model based on $G_b$. Subsequently, the gradient of the potential edge between a fraudster $u_i$ and the target news $q_k$ is calculated as: 
\begin{equation}
\begin{gathered}
\nabla_{a_{q_k,u_i}} = \frac{\partial L(f_{\theta_b}(A_{G_b},X_{G_b}))_{q_k}}{\partial {a_{q_k,u_i}}},\\
\end{gathered}
\label{eq_gradb}
\end{equation}
where $A_{G_b}$ is the adjacency matrix of $G_b$, $X_{G_b}$ is the feature matrix of $G_b$ and $L$ is the cross-entropy loss function. In general, a larger gradient of the potential edge indicates that the corresponding user's participation in the attack will cause a greater impact on the target news. Hence, we select the fraudster $u_i$ with the largest gradient in those remaining fraudsters and update the selected fraudster set $U_a = U_a \cup \{u_i\}$. The related edge is added to $E_b$ and the selection process will be repeated until $\Delta$ fraudsters have been chosen.

\subsection{Post Injection Module}
Once a set of users $U_a$ engaging in the attack is selected, we attempt to determine the targets of sharing behavior for these users such that the shared relations can be established. 
\paragraph{Post Initialization.}
Since the budget for new interactions and the number of selected users are both $\Delta$, each user $u_i \in U_a$ can send a new post to maximize user participation within the attack budget. Hence, we can obtain a set of new post $P^\prime$ and a set of user-post relation $E^\prime_u$. Note that the content of the post is still hard to generate and we temporarily set it to empty. Intuitively, directly sharing the target news $q_k$ is not always the best choice. To measure the impact of selecting different shared targets, we employ the previous propagation tree $G^k_t$ to model the existing propagation structure of the news $q_k$. Then, the sharing behaviors can be treated as injecting posts into $G^k_t$ such that we obtain a new propagation tree $G_o=(\mathcal{V}_o, E_o)$, where $\mathcal{V}_o=\mathcal{V}^k_t \cup P^\prime$ and $E_o = E^k_t$. The new adjacency matrix of $A_{G_o}$ can be formulated as:
\begin{equation}
A_{G_o}= \begin{bmatrix}
A_{G^k_t} & B\\
B^{T} & O\\
\end{bmatrix},   
\label{eq_mt}
\end{equation}
where $A_{G^k_t}$ is the original adjacency matrix of $G^k_t$, $B$ is the potential edges between injected posts and the posts in the original tree and $O$ is the potential edges between injected posts. At the beginning, $B$ and $O$ are both zero matrices.
\paragraph{Connection Optimization.}
Subsequently, we leverage the gradient of the edges in $B$ as the proxy of attention to find the optimal connection. Since the content of the post has not been optimized, it is infeasible to optimize the connection based on the text representation. Instead, we utilize the user representation to select the shared target. According to $E_u$ and $E^\prime_u$, we let the representations of post nodes in $G_o$ equal to the user representation of the post publisher. Then, a GNN-based graph classifier $f_{\theta_t}$ is utilized as a surrogate model to obtain gradient information. Different from the previous process, we only consider $\Delta$ injected posts. Hence, the simplified optimization process is described as:
\begin{equation}
\begin{gathered}
\mathop{\arg\max}\limits_{a_{ij}}\nabla_{B}\odot C = \mathop{\arg\max}\limits_{a_{ij}}\frac{\partial L(f_{\theta_t}(A_{G_o},X_{G_o}))}{\partial B} \odot C,\\
\end{gathered}
\label{eq_gradt}
\end{equation}
where $\odot$ represents the element-wise product, $A_{G_o}$ is the adjacency matrix of $G_o$, $X_{G_o}$ is the user representation matrix of $G_o$, $C$ is a mask matrix and $a_{ij}$ is the edge with the largest gradient under constraints in $B$. At the beginning, $C$ is a matrix of ones. We treat $a_{ij}$ represents an optimal shared relation and expand $E_o=E_o \cup \{(v_i, v_j)\}$. Each post has a unique source in the tree. To ensure each new post can only be selected once, we let $C_{kj}=0, \forall k$. Iteratively, we expand $E_o$ until each injected post has one connection in $E_o$. Hence, we obtain the set of new shared relation $E^\prime_t=E_o \setminus E^k_t$.

\paragraph{Content Clone.}
According to the new propagation tree $G_o$, we optimize the text representation for injected posts which has been set as empty before. Recent study \cite{Wang:brother2023} directly set the feature as zero, indicating an empty message, leading to little impact on detectors. To address this problem, we attempt to clone existing content into our new posts. Similarly, we leverage the model's attention on the text representations to estimate the influence of the existing content. Specifically, we extract text representations from the content of existing posts and retrain a GNN-based graph classifier $f^\prime_{\theta_t}$ according to the text representations. Then, for a new post $p_i \in P^\prime$, we let the feature vector of it equal to the zero feature vector due to its empty content. Given the modified propagation tree $G_o$, the attention weight of the text representation for the new post $p_i$ can be calculated as: 
\begin{equation}
\begin{gathered}
w_{p_i} = \frac{\partial f^\prime_{\theta_t}(A_{G_o},X^\prime_{G_o})_{1-y_{q_k}}}{\partial x_{p_i}},\\
\end{gathered}
\label{eq_wp}
\end{equation}
where $X^\prime_{G_o}$ represents the text representation matrix of $G_o$, $x_{p_i}$ is the feature vector for post $p_i$. Then, we randomly sample a set of potential post $P_r \subset P$. For each $p_j \in P_r$, we calculate the score as:
\begin{equation}
\begin{gathered}
s_{p_{i,j}} = x_{p_j} \cdot w_{p_i},\\
\end{gathered}
\label{eq_sp}
\end{equation}
where $x_{p_j}$ is the text representation of the post $p_j$. After computing the score for all posts in $P_r$, we select the potential post that has the largest score and clone the post content to the new post $p_i$. The content of all injected posts will be updated and then the posts can be injected into the social context. 

Based on the two modules, we mix the generated records of sharing behavior into the raw data and then fool the GNN-based fake news detectors.

\section{Experiment}

\subsection{Experimental Settings}


\paragraph{Datasets.}
We adopt two real-world datasets \cite{Shu:fnn2017,Fey:pyg2019}, i.e., Politifact and Gossipcop, from the PyTorch-Geometric library. 
To train detectors and our surrogate model, we split the data into 20\% for the training, 10\% for the validation, and 70\% for the testing. The testing set is also treated as the set of the news to be attacked.

\begin{table*}[tb]

\centering

\scalebox{0.8}{
\begin{tabular}{lllcccccccccccc}
\toprule
\multirow{2.5}{*}{Dataset}    & \multirow{2.5}{*}{Graph} & \multirow{2.5}{*}{Model} & \multicolumn{6}{c}{Fake News}                                & \multicolumn{6}{c}{Real News}                                \\ \cmidrule(lr){4-9} \cmidrule(lr){10-15} 
                            &                        &                        & - & Random & DICE & MARL & SGA           & GAFSI          & - & Random & DICE & MARL & SGA           & GAFSI          \\ \midrule
\multirow{9}{*}{Politifact} & \multirow{3}{*}{G1}    & GCN                    & 0.20  & 0.16   & 0.29 & 0.97 & \textbf{1.00} & \textbf{1.00} & 0.04  & 0.18   & 0.30 & 0.81 & \textbf{0.95} & \textbf{0.95} \\
                            &                        & SAGE                   & 0.27  & 0.30   & 0.37 & 0.56 & 0.84          & \textbf{0.85} & 0.12  & 0.12   & 0.15 & 0.32 & 0.65          & \textbf{0.70} \\
                            &                        & GAT                    & 0.21  & 0.24   & 0.34 & 0.43 & 0.68          & \textbf{0.70} & 0.10  & 0.12   & 0.18 & 0.39 & \textbf{0.67} & 0.55          \\ \cmidrule(l){2-15} 
                            & G2                     & DECOR-GCN              & 0.18  & 0.31   & 0.45 & 0.57 & 0.86          & \textbf{0.92} & 0.10  & 0.11   & 0.21 & 0.36 & 0.32          & \textbf{0.37} \\ \cmidrule(l){2-15} 
                            & \multirow{3}{*}{G3}    & UPFD-GCN               & 0.26  & 0.29   & 0.31 & 0.39 & 0.55          & \textbf{0.98} & 0.10  & 0.11   & 0.11 & 0.13 & 0.15          & \textbf{0.72} \\
                            &                        & UPFD-SAGE              & 0.26  & 0.35   & 0.52 & 0.56 & 0.89          & \textbf{0.96} & 0.13  & 0.16   & 0.26 & 0.32 & 0.73          & \textbf{0.86} \\
                            &                        & UPFD-GAT               & 0.27  & 0.26   & 0.28 & 0.31 & 0.45          & \textbf{0.48} & 0.14  & 0.16   & 0.17 & 0.25 & 0.41          & \textbf{0.52} \\ \cmidrule(l){2-15} 
                            & G4                     & BiGCN                  & 0.21  & 0.22   & 0.23 & 0.29 & 0.31          & \textbf{0.91} & 0.10  & 0.12   & 0.15 & 0.13 & 0.18          & \textbf{0.92} \\ \midrule
\multirow{9}{*}{Gossipcop}  & \multirow{3}{*}{G1}    & GCN                    & 0.02  & 0.01   & 0.04 & 0.63 & \textbf{0.77} & 0.57          & 0.08  & 0.15   & 0.18 & 0.75 & \textbf{1.00} & \textbf{1.00} \\
                            &                        & SAGE                   & 0.10  & 0.09   & 0.10 & 0.31 & 0.37          & \textbf{0.92} & 0.02  & 0.12   & 0.13 & 0.35 & 0.85          & \textbf{0.91} \\
                            &                        & GAT                    & 0.04  & 0.05   & 0.08 & 0.21 & \textbf{0.40} & 0.23          & 0.03  & 0.08   & 0.11 & 0.22 & \textbf{0.48} & \textbf{0.48} \\ \cmidrule(l){2-15} 
                            & G2                     & DECOR-GCN              & 0.07  & 0.06   & 0.09 & 0.23 & 0.64          & \textbf{0.78} & 0.04  & 0.13   & 0.15 & 0.08 & 0.20          & \textbf{0.24} \\ \cmidrule(l){2-15} 
                            & \multirow{3}{*}{G3}    & UPFD-GCN               & 0.02  & 0.02   & 0.04 & 0.05 & 0.06          & \textbf{0.85} & 0.05  & 0.06   & 0.08 & 0.08 & 0.12          & \textbf{0.84} \\
                            &                        & UPFD-SAGE              & 0.02  & 0.03   & 0.06 & 0.04 & 0.21          & \textbf{0.68} & 0.05  & 0.05   & 0.07 & 0.18 & 0.52          & \textbf{0.68} \\
                            &                        & UPFD-GAT               & 0.02  & 0.03   & 0.06 & 0.04 & 0.20          & \textbf{0.75} & 0.06  & 0.06   & 0.08 & 0.16 & 0.44          & \textbf{0.64} \\ \cmidrule(l){2-15} 
                            & G4                     & BiGCN                  & 0.04  & 0.05   & 0.07 & 0.05 & 0.08          & \textbf{0.63} & 0.05  & 0.07   & 0.08 & 0.09 & 0.10          & \textbf{0.78} \\ \bottomrule
\end{tabular}}
\caption{The success rate of GAFSI compared to other baselines. Fake News and Real News represent the original label of the target news. ``-" denotes the misclassification rate for the detectors before the attack. The best results are highlighted in bold.}
\label{tab_suctotal}
\vspace{-2mm}
\end{table*}

\paragraph{GNN-based Fake News Detectors.}
We investigate the robustness of different GNN-based fake news detectors. For social-context-based detectors, following~\cite{Wang:marl2023}, we construct a user-news bipartite graph. Leveraging Glove~\cite{Pennington:glove2014}, we extract the features of user nodes from their historical posts and the features of news nodes from the news contents. Subsequently, we attack three variants of GNN, i.e., GCN~\cite{Kipf:gcn2017}, GraphSAGE~\cite{Hamilton:sage2017}, and GAT~\cite{Veličković:gat2018}. Besides, following~\cite{Wu:decor2023}, we construct a news engagement graph and attack the related proposed model. For propagation-based detectors, following~\cite{Dou:upfd2021}, we construct a tree-structured propagation graph. The features of post nodes are derived from the historical posts of their owner. Then, we investigate three variants of UPFD ~\cite{Dou:upfd2021}. We also investigate BiGCN \cite{Bian:bigcn2020} on a propagation and dispersion bi-directional tree. Different from the previous tree, the features of post nodes are derived from the post content. 

For clarity, we use G1, G2, G3, and G4 to represent a bipartite graph, a news engagement graph, a propagation tree, and a bi-directional tree mentioned above, respectively.
\paragraph{Baselines.}
Since there is no attempt to conduct a general attack on detectors based on different graph structures, we need to make some necessary assumptions to extend the existing methods. The baseline method will conduct the attack on the user-news bipartite graph and connect users with the target news. To obtain a complete record of social interaction, we assume the users selected by the baselines randomly share the related post of target news and the new posts have empty content. The detailed baselines are described as follows:

\textbf{Random}, which randomly establishes a connection between a user and the target news.

\textbf{DICE}~\cite{Waniek:dice2018}, which randomly connects the target news to several users with a different label. The pseudo-labels of users are obtained from the surrogate model.

\textbf{SGA}~\cite{Li:sga2021}, one of the most effective targeted attacks on the node classification task. SGA constructs a subgraph for the target node and computes gradients of edges in the subgraph to generate edge perturbation.

\textbf{MARL}~\cite{Wang:marl2023}, which is the first work to attack GNN-based fake news detectors utilizing a multiagent reinforcement learning framework.

\begin{figure*}[tb]

    \centering
    \includegraphics[width=7in]{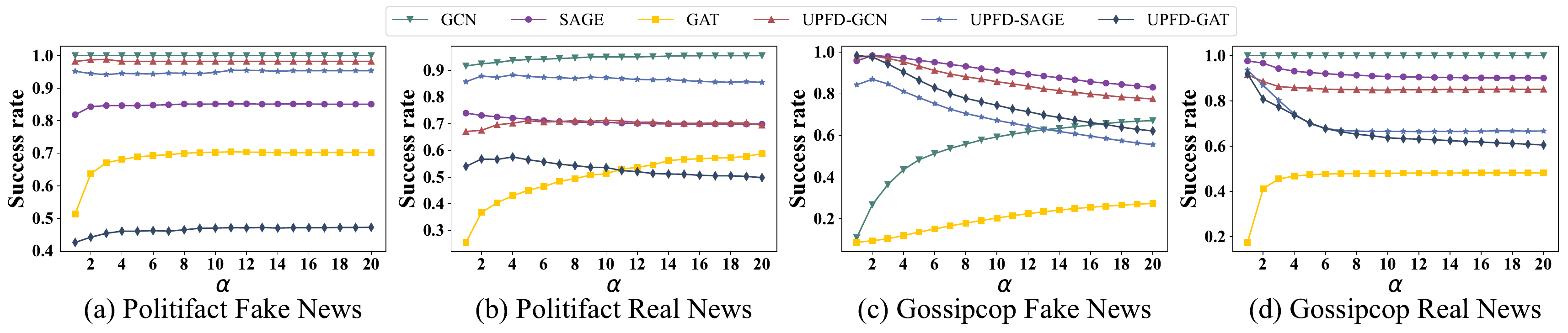}
    \vspace{-6mm}
    \caption{The success rate of GAFSI when adopting different trade-off parameters $\alpha$.}
    \label{fig_alpha}
    \vspace{-3mm}

\end{figure*}

\begin{figure*}[tb]

    \centering
    \includegraphics[width=7in]{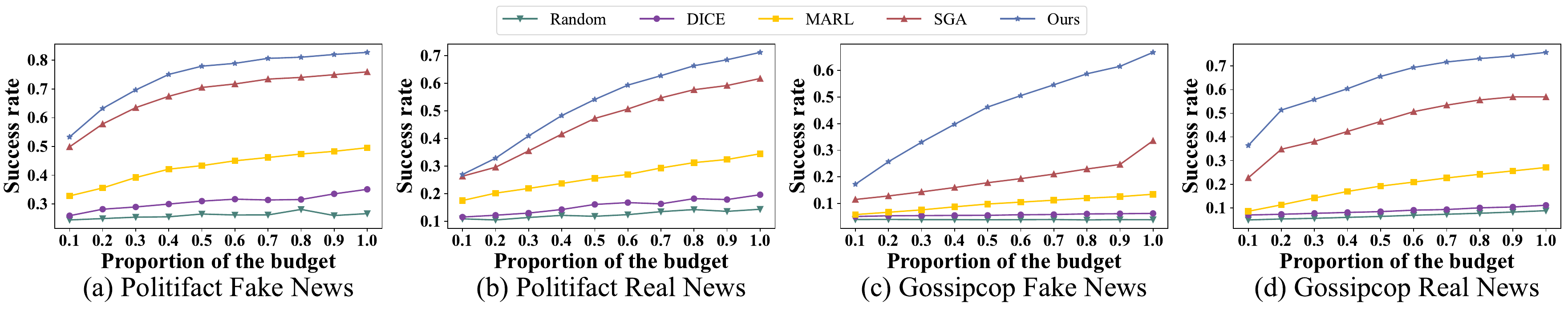}
    \vspace{-6mm}
    \caption{The average success rate of attacks with different budgets.}
    \label{fig_budget}
    \vspace{-3mm}
\end{figure*}

\subsection{Performance Analysis}

\paragraph{Comparison with State-of-the-art Methods.}

Our goal is to change the prediction of the target news. To this end, we evaluate the success rate of the attack which is defined as the number of misclassified news after the attack divided by the total number of target news. A larger success rate means better attack performance. Following the existing targeted attacks~\cite{Zügner:nettack2018,Li:sga2021}, the budget of new records $\Delta$ equals to the degree of the target node in the user-news bipartite graph which represents the original number of people engaged in sharing the news. For each target news, we repeat the attack 10 times and report their average results. 

Table \ref{tab_suctotal} shows the attack performance when changing the prediction of fake news and real news respectively. Our method outperforms all baselines in most cases, particularly against propagation-based detectors on G3 and G4. Besides, GAFSI achieves a greater improved amount of success rate in the larger dataset, i.e., Gossipcop, demonstrating the ability to deal with large-scale graphs. MARL fails to reduce the search space, achieving poor attack performance. Compared to SGA, GAFSI has less influence on target news in three cases. This is because SGA solely focuses on modifying the user-news bipartite graph, while GAFSI makes a trade-off between different graph structures, improving the transferability with relatively minor costs. Furthermore, the attack performance between different variants also improves in some cases, especially against GraphSAGE.

\paragraph{Computational Efficiency.}
We demonstrate the average running time for attacking single news in Table \ref{tab_time}. Compared to SGA, we generate more complex and complete social interaction in a similar running time. MARL suffers from low time efficiency due to the vast search space and multiple samplings. In contrast to MARL, our approach shows a noticeable advantage in computational efficiency through utilizing a hierarchical strategy in the fraudster selection module.

\begin{table}[tb]

\centering

\scalebox{0.8}{
\begin{tabular}{lccc}
\toprule
Dataset    & MARL   & SGA           & GAFSI \\
\midrule
Politifact & 51.77 & 0.78 & \textbf{0.50} \\
Gossipcop  & 54.24 & 2.20 & \textbf{1.69} \\
\bottomrule
\end{tabular}}
\vspace{-1mm}

\caption{The average running time (s) of attacking a piece of news.}
\label{tab_time}
\vspace{-3mm}
\end{table}

\subsection{Ablation Studies and Other Analysis}

\paragraph{Effects of Varying Hyper-parameters.}
We examine the performance of GAFSI with different parameters $\alpha$ to make a trade-off between local influence and global influence in the fraudster selection module. The results against detectors on G1 and G3 are shown in Figure \ref{fig_alpha}, GAFSI achieves a poor performance against detectors on G1 when only considering local influence, i.e., $\alpha=1$. As $\alpha$ increases, we observe that enhancing the attack performance against detectors on G1 sacrifices the attack performance against detectors on G3. When $\alpha$ is smaller, this trade-off is worthwhile. Ultimately, the method's attack performance achieves a balance, indicating that subsequent nodes have a low impact on any models.

\begin{table}[tb]

\centering

\scalebox{0.8}{
\begin{tabular}{llcccc}
\toprule
Dataset                     & Model     & - & Root & Random & Ours          \\
\midrule
\multirow{3}{*}{Politifact} & UPFD-GCN  & 0.17  & 0.79 & 0.37   & \textbf{0.84} \\
                            & UPFD-SAGE & 0.19  & 0.81 & 0.86   & \textbf{0.90} \\
                            & UPFD-GAT  & 0.20  & 0.48 & 0.48   & \textbf{0.50} \\
                            
\midrule
\multirow{3}{*}{Gossipcop}  & UPFD-GCN  & 0.04  & 0.82 & 0.39   & \textbf{0.84} \\
                            & UPFD-SAGE & 0.03  & 0.63 & 0.61   & \textbf{0.68} \\
                            & UPFD-GAT  & 0.04  & 0.56 & 0.62   & \textbf{0.70} \\                            
\bottomrule
\end{tabular}}
\vspace{-1mm}
\caption{The success rate for different post sources.}
\label{tab_abtarget}
\vspace{-3mm}
\end{table}

\paragraph{Effects of Different Post Sources.}
To demonstrate the validity of the post injection module in GAFSI, we conduct ablation experiments on the connection strategy. A source of a post could be a news or a related post. To conduct the ablation study, we provide two strategies for comparison. Fraudsters can choose to directly share the target news, which is regarded as a Root strategy. Another choice for fraudsters is to randomly share a related post, which is regarded as a Random strategy. We conduct experiments against detectors on G3. As shown in Table \ref{tab_abtarget}, we compare these two strategies to ours, indicating that our approach can cause a greater impact on the propagation tree of the target news.

\paragraph{Effects of Different Post Content.}
For the ablation study of post content generation, we directly set the new post as an empty text or randomly assign the existing content to it. In contrast to previous experiments, the detectors to be attacked classify news based on the content of posts. Therefore, we leverage text representation to train the detector on G3 and G4 before the attack. Table \ref{tab_abtext} reports the results of different strategies against detectors that leverage text representation as post nodes' features. It is obvious that our strategy contributes to enhancing the attack performance.

\begin{table}[tb]

\centering

\scalebox{0.8}{
\begin{tabular}{llcccc}
\toprule
Dataset                     & Model     & - & Empty & Random & Ours          \\

\midrule
\multirow{4}{*}{Politifact} & UPFD-GCN  & 0.10  & 0.13  & 0.19   & \textbf{0.95} \\
                            & UPFD-SAGE & 0.11  & 0.13  & 0.14   & \textbf{0.68} \\
                            & UPFD-GAT  & 0.11  & 0.12  & 0.12   & \textbf{0.48} \\
                            & BiGCN  & 0.12  & 0.14  & 0.20   & \textbf{0.91} \\
\midrule
\multirow{4}{*}{Gossipcop}  & UPFD-GCN  & 0.03  & 0.04  & 0.17   & \textbf{0.70} \\
                            & UPFD-SAGE & 0.02  & 0.02  & 0.20   & \textbf{0.27} \\
                            & UPFD-GAT  & 0.02  & 0.02  & 0.25   & \textbf{0.39} \\
                            & BiGCN  & 0.04  & 0.12  & 0.12   & \textbf{0.71} \\
\bottomrule
\end{tabular}}
\caption{The success rate for different post content.}
\label{tab_abtext}
\vspace{-3mm}
\end{table}

\paragraph{Effects of Different Budgets.}
To better evaluate the effectiveness of GAFSI with different budgets, as shown in Figure \ref{fig_budget}, we gradually reduce the budget to observe changes in attack performance on the detectors based on G1 and G3, respectively. As the budget increases, our method widens the gap in attack performance compared to other methods, especially for the fake news in Gossipcop. We also observe a slowing growth rate, which could be attributed to two factors. Firstly, later-selected users often have lower influences than earlier-selected ones, Secondly, the increase in the number of users might dilute the impact of single users.

\begin{figure}[!tb]
    \centering
    \includegraphics[width=0.99\linewidth]{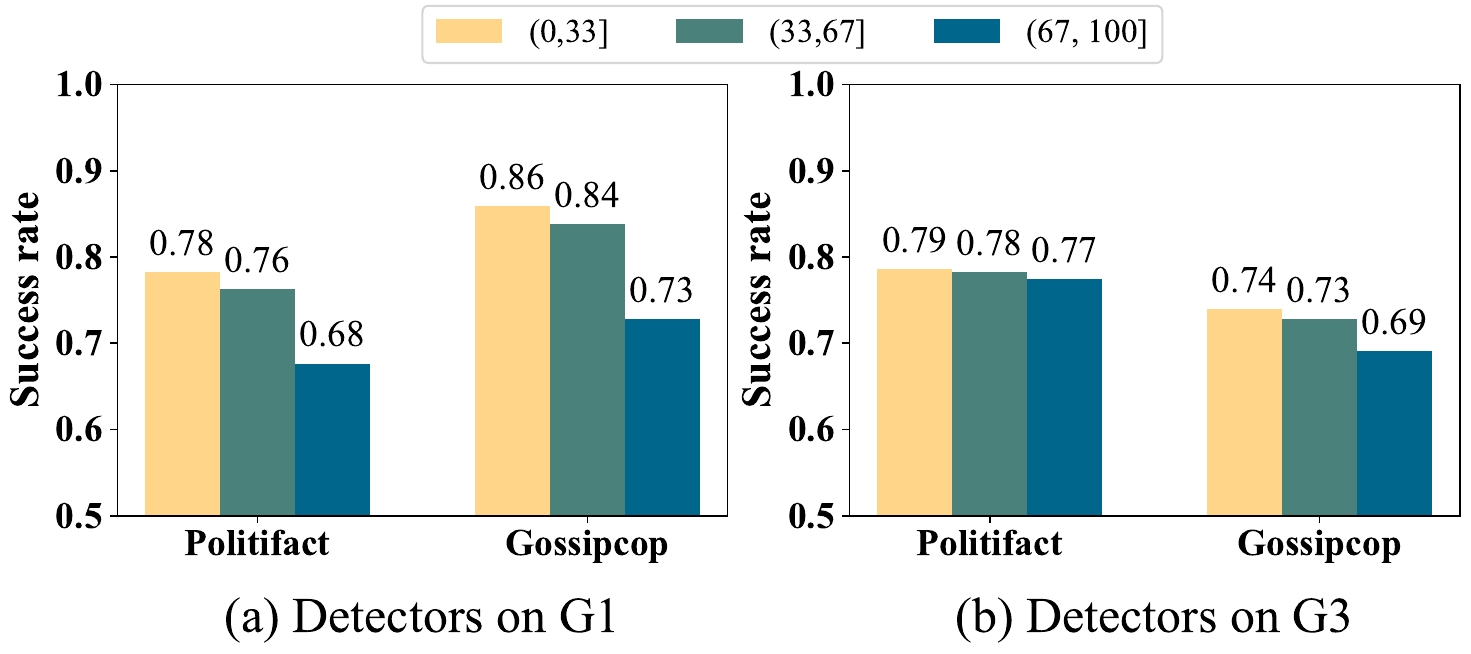}
    \vspace{-2mm}
    \caption{The success rate of GAFSI on target news within different degree ranges.}
    \label{fig_degrees}
    \vspace{-2mm}
\end{figure}

\paragraph{Effects of Different News Degrees.}
Intuitively, the degree of node has a large impact on attack performance. Since relatively low degrees constitute a significant portion of real-world datasets, nodes that have a degree less than 100 are selected for further investigation. We group the target news based on their node degrees on the bipartite graph. Then, we summarize the attack results for the grouped nodes. As shown in Figure \ref{fig_degrees}, the success rate of our method decreases when the degree of target news increases. This implies that the detectors are more robust when predicting news with more participants involved. Besides, compared to detectors working on G1, those working on G3 seem less sensitive facing varying degrees, which may be attributed to their tree structures.


\section{Conclusion}

In this paper, we propose a general black-box adversarial attack framework against GNN-based fake news detectors with different graph structures. The key idea is to add fake social interaction to the social context via fraudster's sharing behaviors. Extensive experimental results demonstrate the effectiveness and superiority of our method in attacking different fake news detectors. In the future, we plan to investigate the engagement pattern and realize a more imperceptible attack.

\section*{Acknowledgements}

This work was supported in part by the National Science Fund for Distinguished Young Scholarship of China (Grant no. 62025602), the National Natural Science Foundation of China (Grant nos. U22B2036, 62073263, 62203363, 62102105), the Fundamental Research Funds for the Central Universities (Grant no. D5000230112), the Tencent Foundation and XPLORER PRIZE, the Young Talent Fund of Association for Science and Technology in Shaanxi (Grant no. 20240105), and the Shaanxi Provincial Natural Science Foundation (Grant no. 2024JC-YBQN-0620).

\bibliographystyle{named}
\bibliography{ijcai24}

\end{document}